\documentclass[10pt,twocolumn,letterpaper]{article}

\usepackage[pagenumbers]{cvpr} 

\usepackage{float}
\usepackage[dvipsnames, table]{xcolor}
\usepackage{duckuments}
\usepackage{bm}

\newcommand{\parnobf}[1]{\vspace{1mm} \par \noindent {\bf {#1}}}

\newcommand{\camready}[1]{\textcolor{black}{#1}}
\definecolor{first}{rgb}{1.0, .83, 0.3}
\definecolor{second}{rgb}{1.0, 0.93, 0.7}
\def \first {\cellcolor{first}\bfseries}
\def \second {\cellcolor{second}}

\newcommand{\ours}[0]{MVD-Fusion}

\definecolor{cvprblue}{rgb}{0.21,0.49,0.74}
\usepackage[pagebackref,breaklinks,colorlinks,citecolor=cvprblue]{hyperref}

\usepackage{tabularx}
\usepackage{multirow}
\usepackage{epigraph}
\usepackage{booktabs}
\usepackage{amsmath}
\usepackage{amssymb}
\usepackage{cuted}

\usepackage{lib}

\def\paperID{479} 
\def\confName{CVPR}
\def\confYear{2024}

\title{\ours: Single-view 3D via Depth-consistent Multi-view Generation}

\author{
Hanzhe Hu$^{1}$\thanks{Equal contribution.}
\and
Zhizhuo Zhou$^{2}$\footnotemark[1]
\and
Varun Jampani$^{3}$
\and
Shubham Tulsiani~$^{1}$
\and 
\\
$^{1}$ Carnegie Mellon University \quad
$^{2}$ Stanford University \quad
$^{3}$ Stability AI \quad\\
\\ {\tt \small \href{https://mvd-fusion.github.io/}{https://mvd-fusion.github.io/}}
}

\begin{document}
\maketitle

\begin{strip}
\vspace{-5em}
\centering
\includegraphics[width=1.\linewidth]{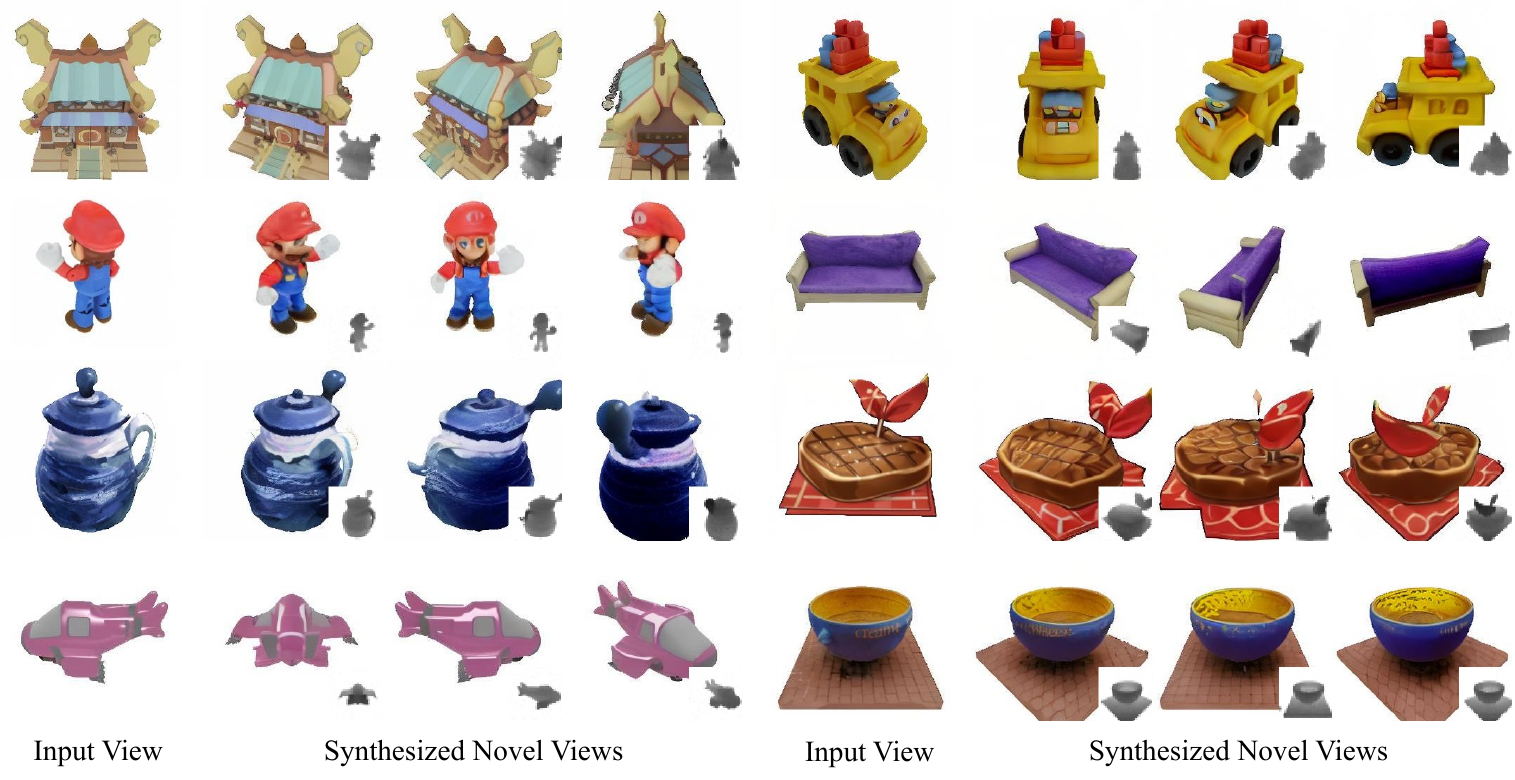}

\captionof{figure}{\textbf{Single-view 3D Inference}. Given an input RGB image, \ours~allows synthesizing multi-view RGB-D images using a depth-guided attention mechanism for enforcing multi-view consistency. We visualize the input RGB image (left) and three synthesized novel views (with generated depth in inset).
}
\figlabel{teaser}
\end{strip}
\begin{abstract}
We present \ours: a method for single-view 3D inference via generative modeling of multi-view-consistent RGB-D images. While recent methods pursuing 3D inference advocate learning novel-view generative models, these generations are not 3D-consistent and require a distillation process to generate a 3D output. We instead cast the task of 3D inference as directly generating mutually-consistent multiple views and build on the insight that additionally inferring depth can provide a mechanism for enforcing this consistency.  Specifically, we train a denoising diffusion model to generate multi-view RGB-D images given a single RGB input image and leverage the (intermediate noisy) depth estimates to obtain reprojection-based conditioning to maintain multi-view consistency. We train our model using large-scale synthetic dataset Obajverse as well as the real-world CO3D dataset comprising of generic camera viewpoints. We demonstrate that our approach can yield more accurate synthesis compared to recent state-of-the-art, including distillation-based 3D inference and prior multi-view generation methods. We also evaluate the geometry induced by our multi-view depth prediction and find that it yields a more accurate representation than other direct 3D inference approaches. 
\end{abstract}
\vspace{-1em}    
\section{Introduction}
\seclabel{sec:intro}
The task of recovering 3D from a single 2D image has witnessed a recent wave of generative-modeling based approaches. In particular, while initial 3D prediction methods pursued inference of volumetric~\cite{choy20163d,girdhar2016learning}, point clouds~\cite{wu2023multiview,fan2016aps}, or meshes~\cite{kanazawa2018learning,gkioxari2019mesh} representations, a class of recent approaches~\cite{zhou2023sparsefusion,liu2023zero1to3,melas2023realfusion} instead formulate the task as learning (conditional) generation of novel views. By adapting large-scale pre-trained generative models, these methods can learn generalizable view synthesis that performs remarkably well even for generic objects in-the-wild. However, the synthesized novel views are not mutually consistent and these 2D generative methods rely on a (costly) `score distillation'~\cite{poole2022dreamfusion} based optimization to then recover a consistent 3D. While this process can yield impressive results, these come at the cost of a reduction in both the efficiency of inference and the diversity of the generations.

In this work, we seek to overcome these limitations, and pursue an approach that allows directly generating diverse  outputs.
We do so by re-formulating the task of 3D inference as that of generating a set of (mutually consistent) multiple views, and learn a (conditional) generative prior to model this joint distribution. 
While some recent (and concurrent) methods do similarly `co-generate' multiple views given a single input image~\cite{szymanowicz2023viewset,shi2023mvdream}, these are typically not geometrically consistent. Instead, our approach is inspired by  the recent work from Liu \etal ~\cite{liu2023syncdreamer} which incorporates a 3D bottleneck with unprojection and reprojection as an inductive bias for ensuring geometric consistency across views. In this work, we explore an alternate mechanism for enforcing such consistency.
In particular, we formulate a depth-generation-guided approach that: a) allows improved generation via depth-based reprojection, and b) enables directly producing an estimate of the 3D geometry via the inferred multi-view 2.5D representation. 

Our approach for enforcing multi-view consistency stems from a simple question: \emph{what does it mean for images to be 3D-consistent?} Drawing inspiration from classical 3D reconstruction methods, one answer to this question is that if a pixel in one image corresponds to a point that is also visible in another, then the local appearance should match. However, how can we know where the 3D point corresponding to a pixel in one image may project in the other? This inspires our solution for generating multi-view consistent images, where we not only generate the RGB images but also reason about the corresponding depth for these generations (and thus allow such reprojection-based multi-view consistency). More specifically, we adopt existing 2D diffusion models to generate RGB-D images while adding multi-view projection based on (noisy) depth estimates to enforce 3D consistency. 

We train our system using a large-scale synthetic dataset, and empirically demonstrate the efficacy of our approach on held out objects as well as scanned real world data. We show that our approach allows more accurate generation compared to prior state of the art while also (directly) generating plausible geometry via the synthesized depth images. Finally, we also highlight the ability of our method to sample diverse outputs and its ability to generalize zero-shot to in-the-wild generic objects.




\section{Related Work}
\seclabel{sec:related}

\paragraph{Single-view 3D Prediction.}
The task of inferring 3D from 2D images is a long-standing one in computer vision, and the pre-dominant learning-based approach has been to frame it as a \emph{prediction} task where a data-driven predictor is trained to output a 3D representation given image input. In particular, several deep learning based methods pursued this task by inferring a plethora of 3D representations such as volumetric 3D~\cite{girdhar2016learning,choy20163d,tulsiani2017multi,ye2021shelf}, meshes~\cite{gkioxari2019mesh,kanazawa2018learning,kulkarni2022directed,kulkarni2023learning,xu2019disn,wang2018pixel2mesh}, point clouds~\cite{wu2023multiview, fan2016aps, navaneet2020ssl3drecon}, or neural implicit fields~\cite{lin2020sdfsrn,mescheder2019occupancy,alwala2022pre}.
While these approaches have shown promising results, they often struggle to generate detailed outputs for complex objects owing to the ambiguity in unobserved regions. Indeed, any such regression-based approach fundamentally cannot model the inherent uncertainty in single-view reconstruction. In contrast, our generative modeling-based approach can synthesize and generate high-fidelity outputs and also yield multiple modes.


\begin{figure*}
    \centering
    \includegraphics[width=1\linewidth]{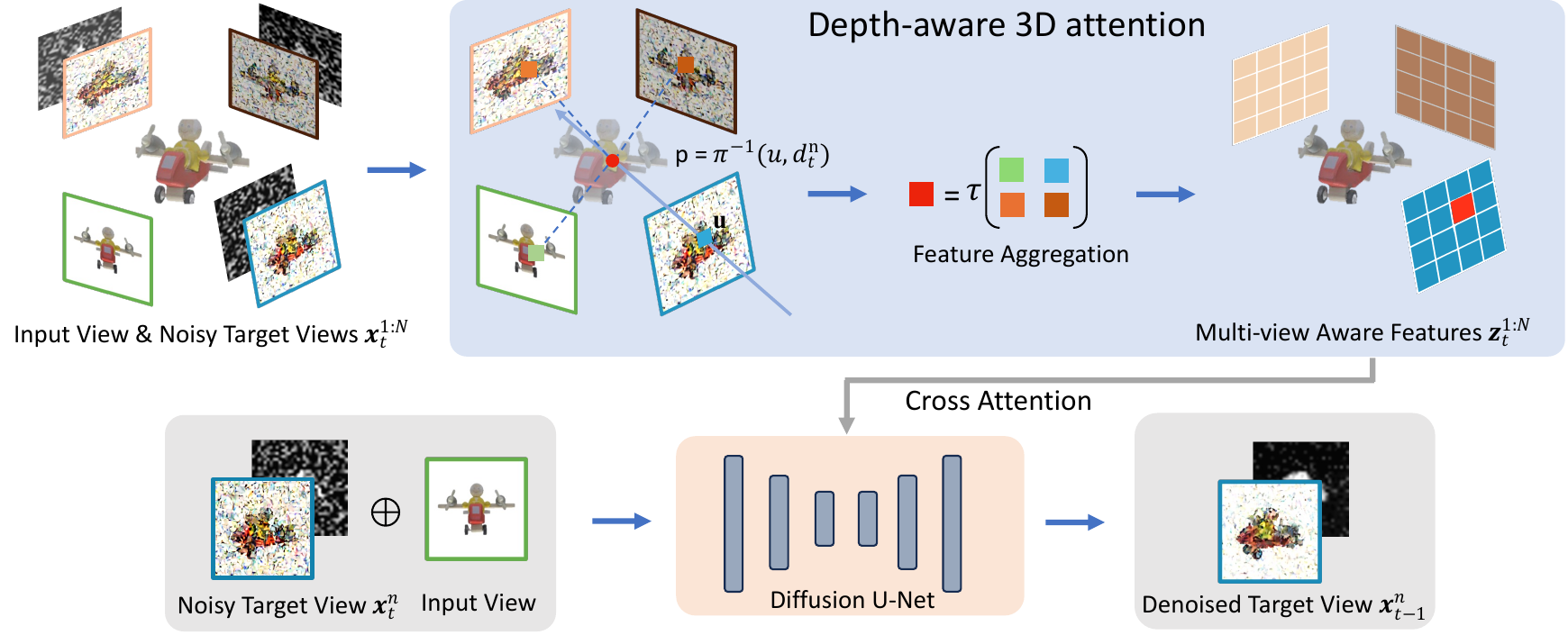}
    \caption{\textbf{Approach Overview.} \ours~learns a denoising diffusion model for generating multi-view RGB-D images given an input RGB image. At each diffusion timestep $t$, \ours~uses the current (noisy) depth estimates to compute depth-projection-based multi-view aware features (top). A novel-view diffusion based U-Net  is modified to leverage these multi-view aware features as additional conditioning while producing denoised estimates of both, RGB and depth (bottom).
    }
    \figlabel{method}
    \vspace{-1.5em}
\end{figure*}

\vspace{-2mm}
\paragraph{3D Inference via 2D Diffusion.}
Instead of directly predicting 3D shapes in a feed-forward way, this line of work utilizes 2D diffusion prior to facilitating 3D inference. In particular, DreameFusion~\cite{poole2022dreamfusion} and SJC~\cite{wang2022score} formulated a `score distillation' objective that enabled incorporating pre-trained diffusion as a prior for optimization, and leveraged it to distill a 2D stable diffusion model for the text-to-3D inference.
Inspired by this, several  works~\cite{melas2023realfusion,qian2023magic123,xu2023neurallift,shen2023anything,tang2023make,deng2023nerdi,gu2023nerfdiff} adopt this pipeline to optimize a neural radiance field (NeRF~\cite{mildenhall2020nerf}) for the single-view reconstruction task. For instance, RealFusion~\cite{melas2023realfusion} extracts from a single image of an object a 360$^\circ$ 3D reconstruction by leveraging a 2D diffusion model with a single-image variant of textual inversion whereas NeuralLift-360~\cite{xu2023neurallift} utilizes a depth-aware NeRF and learns to craft the scene by denoising diffusion models. However, as these methods only use pre-trained image diffusion models for 3D inference, they can suffer from implausible 3D outputs \eg \emph{janus effect} and do not always preserve the details in the observed image.

To circumvent this, SparseFusion~\cite{zhou2023sparsefusion} proposed to learn a novel-view diffusion model using epipolar feature transformer to build a view-conditioned features model for sparse-view reconstruction and distilled it to obtain more accurate 3D reconstructions. Moreover, Diffusion with Forward Models \cite{tewari2023forwarddiffusion} applied 2D diffusion networks to denoise a consistent 3D scene. Our approach builds on Zero-1-to-3~\cite{liu2023zero1to3}, which demonstrated that a large pre-trained image diffusion model can be finetuned for novel-view generation using a large-scale 3D dataset to achieve better generalization ability. While these methods are able to produce high-quality predictions, the reliance on score distillation sampling restricts them from obtaining diverse results with single-view 3D prediction as an under-constrained task.

\vspace{-2mm}
\paragraph{Multi-view Image Generation.}
Unlike novel-view generation models which model the distribution over a single view given a reference image,  many recent works have investigated generating multi-view images at the same time by using diffusion models, including text-based methods~\cite{tang2023mvdiffusion,shi2023mvdream} and image-based methods~\cite{szymanowicz2023viewset,liu2023syncdreamer}. Given a text conditioning, MVDiffusion~\cite{tang2023mvdiffusion} simultaneously generates all images with a global transformer to facilitate cross-view interactions. Similarly, MVDream~\cite{shi2023mvdream} produces multi-view images via multi-view diffusion and leverages a self-attention layer to learn cross-view dependency and encourage multi-view consistency. While these methods rely on text input, Viewset Diffusion~\cite{szymanowicz2023viewset} adopts a similar approach for generating a multi-view image set given an input image and subsequently infers a radiance field to ensure consistent geometry. While these methods, similar to our goal, can model the distribution over novel views, they do leverage any geometric mechanism to enforce multi-view consistency. Perhaps most closely related to our work, SyncDreamer~\cite{liu2023syncdreamer} proposes to use a 3D-aware feature attention mechanism that correlates the corresponding features across views to enforce multi-view consistency. Different from ~\cite{liu2023syncdreamer}, we utilize depth information to learn consistency across views instead of a 3D bottleneck that contains redundant information. \camready{Finally, there have been several promising concurrent works which also pursue multi-view inference \cite{shi2023zero123plus, liu2023one2345, kant2024spad, long2023wonder3d, wang2024crm, huang2023epidiff} but we believe that our method of depth-guided multi-view diffusion represents a complementary advance to the techniques proposed in these.}


\section{Method}
\seclabel{method}

Given a single RGB image, our method generates a set of multi-view consistent RGB-D predictions. In addition to allowing the synthesis of the object from any desired set of views, the generated multi-view depth maps also conveniently yield a (coarse) point cloud representation of the geometry. To ensure multi-view consistency among the generated images, we model the joint distribution over a set of posed images by adding depth-guided 3D cross-attention layers on top of pre-trained latent diffusion backbone from Stable Diffusion \cite{rombach2022high} and Zero-1-to-3 \cite{liu2023zero1to3}. We first formalize this task of multi-view generation via denoising diffusion (\secref{method_multi}), and then detail our specific approach to enforcing multi-view consistency (\secref{method_consistency}) and  2.5D image generation (\secref{method_rgbd}). A diagram of our method is in \figref{method}.

\subsection{Multi-view Denoising Diffusion}
\seclabel{method_multi}
A conditional denoising diffusion model can capture the distribution over a variable of interest $\mathbf{x}$ given some conditioning $\mathbf{c}$. In particular, by learning a function $\epsilon(\mathbf{x}_t, \mathbf{c}, t)$ that learns to denoise an input with time-dependent corruption,  diffusion models can allow sampling from the distribution $p(\mathbf{x}|\mathbf{c})$. Towards our goal of multi-view generation, we are interested in an instantiation of this framework where the conditioning corresponds to an observed RGB image $\mathbf{y}$ and a set of desired novel viewpoints $\{\pi^n\}$. Given these as input, we aim to generate a (mutually consistent) set of novel views $\{\mathbf{x}^n\}$ corresponding to the conditioning viewpoints and thus seek to learn a denoising diffusion model that captures $p(\{\mathbf{x}^n\} | \mathbf{y}, \{\pi^n\})$. 

To learn such a diffusion model, we need to formulate an approach that can predict the noise added to a set of corrupted multi-view images: 
\begin{gather*}
    \{ \mathbf{x}^n_t \} = \{ \sqrt{\bar{\alpha}_t} \mathbf{x}^n + \sqrt{1-\bar{\alpha}_t} \epsilon_n \} \\
    \epsilon_{pred} = f(\{ \mathbf{x}^n_t \}, \mathbf{y}, t)
\end{gather*}
Instead of learning such a prediction model from scratch, we propose to adapt a pre-trained novel-view generative model from Zero-1-to-3~\cite{liu2023zero1to3}. Specifically, this model captures the distribution over a single novel view $p(\mathbf{x}|\mathbf{y}, \mathbf{\pi})$ given an RGB input by learning a denoising function $\epsilon_{\phi}(\mathbf{y}, \mathbf{x}_t, \pi, t)$. While we aim to leverage this pre-trained large-scale module for efficient learning and generalization, it only models the distribution over a single novel view whereas we aim to model the joint distribution over multiple views.  To enable efficiently adapting this, we propose a first learn a separate module that computes view-aligned multi-view aware features $\{\mathbf{z}^n_t\}$. We then modify the pre-trained single-view diffusion model to additionally leverage this multi-view aware conditioning:
\begin{gather}
    \mathbf{z}^n_t = {f_{\theta}(\mathbf{y}, \{\mathbf{x}_t^n\}, \{\pi^n\}, t)} \\
    \epsilon_{pred} = \{\epsilon_{\phi'}(\mathbf{y}, \mathbf{x}_t^n, \mathbf{\pi}^n, \mathbf{z}_t^n, t) \}
\end{gather}

\begin{figure}
    \centering
    \includegraphics[width=1\linewidth]{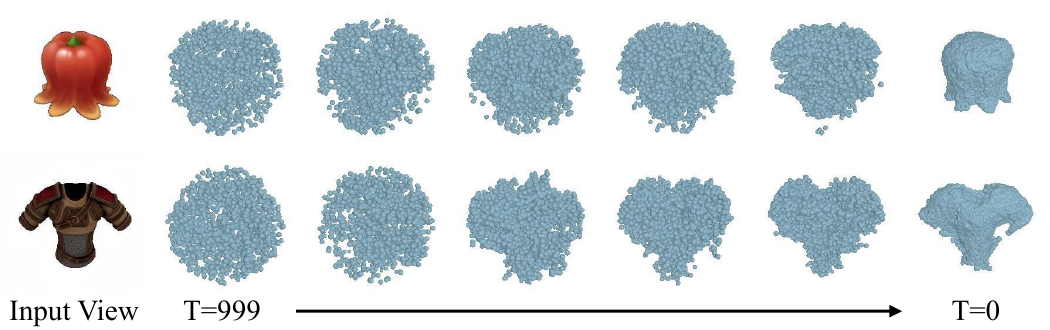}
    \caption{We visualize the unprojected point cloud obtained from a set of noisy RGB-D images at different timesteps during inference. We observe the gradual denoising of geometry from a random point cloud to a point cloud that matches the input object. }
    \figlabel{pc-diffusion}
\end{figure}
\subsection{Depth-guided Multi-view Consistency}
\seclabel{method_consistency}
Generating a set of consistent images requires the network to attend across different images within the set. SyncDreamer \cite{liu2023syncdreamer} proposes a way of achieving multi-view attention by unprojecting features from the set of images $\{\mathbf{x}^{0}_t,\mathbf{x}^1_t,...\mathbf{x}^N_t\}$ onto a 3D volume $\mathcal{V}$ and interpolating conditioning feature frustums $\{\mathbf{z}^{0}_t,\mathbf{z}^1_t,...\mathbf{z}^N_t\}$. However, interpolating feature frustums linearly spaced across the whole 3D volume is an expensive operation that assumes no prior knowledge of object surfaces. In contrast, we propose to explicitly reason about the surface by additionally generating depth and biasing sampling near the possible surface.


Given a target view $\mathbf{x}^i_t$, we obtain feature frustum $\mathbf{z}^i_t$ by shooting rays and sampling features at 3D locations along the rays. For each ray, we sample $D$ depth values near the expected surface and aggregate projected features from target views $\{\mathbf{x}_t^n\}$ and input view $\mathbf{y}$. Let $(\mathbf{z}^i_t)_{md}$ be the feature for the $m$-th ray at $d$-th depth in $\mathbf{z}^i_t$. For a 3D point $\mathbf{p}^i_{md}$ corresponding to the feature $(\mathbf{z}^i_t)_{md}$, we sample $N+1$ features $\mathbf{c}_{mdn}$ from $\{\mathbf{x}_t^n\}$ and $\mathbf{y}$. We also include the plucker embedding of query ray $\mathbf{q}_m$ and reference rays $\mathbf{r}_{mdn}$ from $\mathbf{p}^i_{md}$ to $N+1$ camera centers along with the sampled features as input into the transformer $f_\theta$ that predicts view-aligned multi-view aware features: 
\begin{equation} \eqlabel{zti}
\begin{split}
(\mathbf{z}^i_t)_{md} = \sum_{n=0}^{N+1} w_\theta(\mathbf{v}_{mdn},  t) f_\theta(\mathbf{v}_{mdn}, t)
\\ \text{where}~\mathbf{v}_{mdn} = \{\mathbf{c}_{mdn}, \mathbf{r}_{mdn}, \mathbf{q}_{m}\}
\end{split}
\end{equation}
Here, $w_\theta(\mathbf{v}_{mdn},  t)$ represents (normalized) weights predicted by the transformer, which are then used to aggregate the multi-view features to obtain the pointwise feature $(\mathbf{z}^i_t)_{md}$.

Naively, we can sample a large number of depth points along each ray linearly spaced throughout the scene bound; however, such exhaustive sampling quickly becomes a memory constraint while possibly making the learning task more difficult as the network may also observe features away from the surface. Thus, we sample $D=3$ depths from a Gaussian distribution centered around an unbiased estimate of depth given the noisy depth $d_t$ and a scaled version of the denoising diffusion model variance schedule in equation \eqref{depth_estimate}.
\begin{equation} \eqlabel{depth_estimate}
\begin{split}
d \sim \mathcal{N}(\mathbb{E}\left[d_0\right], k\frac{\sqrt{\bar{\alpha_t}}}{\sqrt{1-\bar{\alpha_t}}})
\\ \text{where}~d_t = \sqrt{\bar{\alpha}_t} d_0 + \sqrt{1-\bar{\alpha}_t} \epsilon 
\\ \text{and}~\mathbb{E}\left[d_0\right] = \frac{d_t}{\sqrt{\bar{\alpha_t}}}
\end{split}
\end{equation}

We then use these multi-view aware features $\{\bm{z}^n_t\}$ as conditioning input into our latent diffusion model in \secref{method_rgbd}.


\begin{figure*}[!htbp]
    \centering
    \includegraphics[width=0.99\linewidth]{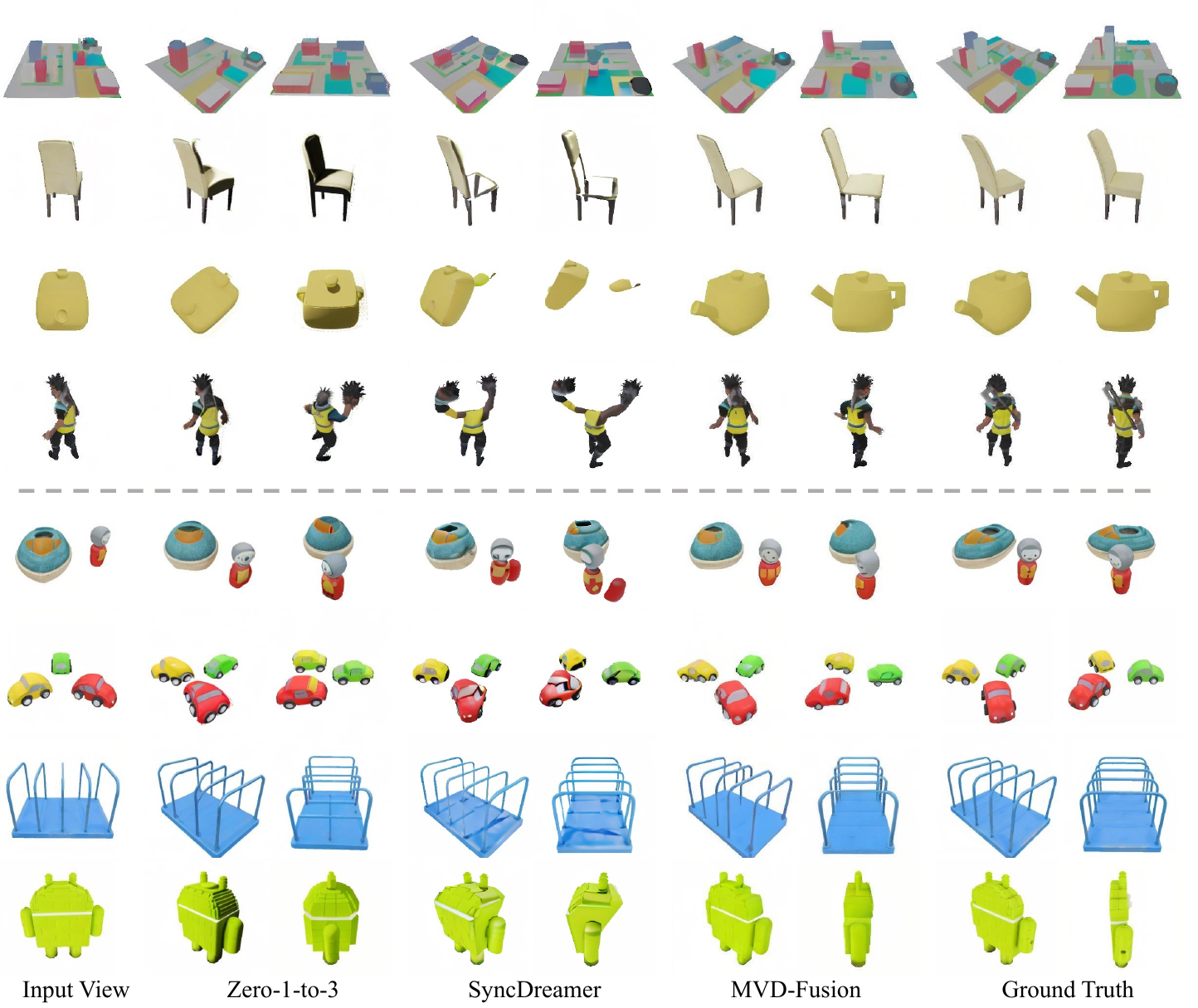}
    \caption{\textbf{Qualitative results for novel view synthesis on instances from Objaverse (top) and Google Scanned objects (bottom)}. We compare our method with Zero-1-to-3~\cite{liu2023zero1to3} and SyncDreamer~\cite{liu2023syncdreamer}. We show the input image and two novel views generated by each method. Zero-1-to-3 independently generates novel views which are not consistent (\eg the person in Objaverse). While both, SyncDreamer and \ours~yield consistent generations, we find that \ours~can generate more plausible output (\eg the Android image) and is more faithful to details in the input (\eg the three cars).} 
    \figlabel{nvs-obja-gso-fig}
    \vspace{-0.5em}
\end{figure*}

\subsection{Learning Multi-view 2.5D Diffusion}
\seclabel{method_rgbd}
Inspired by the success of finetuning pretrained Stable Diffusion models, we adapt Zero-1-to-3 \cite{liu2023zero1to3} as our multi-view novel view synthesis backbone $\epsilon_\phi$. While Zero-1-to-3 is designed to only model single-view distributions and generate RGB output, we adapt it to predict an additional depth channel and cross-attend to the multi-view aware features.

First, we increase the input and output channels of the latent diffusion UNet backbone to predict normalized depth. While the image latents can be decoded into high-resolution images, our predicted depth map remains at the lower resolution. This multi-resolution approach to predicting RGB-D lets us use the frozen Stable Diffusion VAE to decode high-resolution RGB images. Moreover, we add additional residual cross-attention layers at multiple levels of the UNet to attend to our multi-view aware features. \camready{Finally, we modify the camera parameterization used in Zero-1-to-3 from a 3-DoF azimuth, elevation, and radius parameterization to use the full perspective camera matrix. This makes our method capable of handling arbitrary camera poses in real datasets such as CO3D.}

During training, we finetune all the parameters of our network and follow \cite{ho2020denoising} to use a simplified variational lower bound objective in
\eqref{diffusion}. During inference, we follow \cite{liu2023syncdreamer} and use a classifier free guidance of 2.0. 

\begin{equation} \eqlabel{diffusion}
\begin{split}
\mathcal{L}_{DM} = \mathbb{E}_{{\mathbf{x}}_0^n, \mathbf{\epsilon}^n, t}\left[||\mathbf{\epsilon}^n - \{\epsilon_{\phi'}(\mathbf{y}, \mathbf{x}_t^n, \mathbf{\pi}^n, \mathbf{z}_t^n, t) \}||\right]
\\ \text{where}~\mathbf{x}_t = \sqrt{\bar{\alpha_t}} \mathbf{x}_0 + \sqrt{1-\bar{\alpha_t}} \mathbf{\epsilon}; ~~ \mathbf{\epsilon} \sim \mathcal{N}(0,1)
\end{split}
\end{equation}

\section{Experiments}
We train \ours~ using a large-scale synthetic dataset (\secref{exp_setup}), and evaluate it on both, synthetic and real-world objects for view synthesis (\secref{exp_obja}) and 3D reconstruction (\secref{exp_3d}). We show that our results achieves more accurate view synthesis compared to state-of-the-art as well as yields better 3D predictions compared to prior direct 3D inference methods. Finally, we present qualitative results on in-the-wild-objects (\secref{exp_add}).

\subsection{Experimental Setup}
\seclabel{exp_setup}

\paragraph{Datasets.}
We use the large-scale 3D dataset Objaverse~\cite{deitke2023objaverse} for training. Since Objaverse is large (800K instances) and contains several instances with poor texture, we filter the dataset with CLIP score to remove instances that match a set of hand-crafted negative text prompts. Our filtered set contains around 400K instances. For each instance, we render 16 views from an elevation of 30 degrees and azimuth linearly spaced across 360 degrees. Additionally, we hold out a subset of Objaverse instances following \cite{liu2023zero1to3} for evaluation, which consists of about 4k instances. 

Beyond evaluating on these held-out Objaverse instances, we also evaluate our method with the Google Scanned Object dataset (GSO)~\cite{downs2022google}, which consists of high-quality scanned household items. For each object, we render 16 views evenly spaced in azimuth from an elevation of 30 degrees and choose one of them as the input image. For quantitative results, we randomly chose 30 objects to compute the metrics. \camready{Finally, to show the flexibility of our approach in modeling real-world datasets with general perspective cameras, as opposed to the common 3DoF cameras used in Objaverse and GSO, we finetune and evaluate our model on CO3D~\cite{reizenstein2021common}. We follow \cite{zhang2022relpose} to train on 41 categories and evaluate on the held-out set of all 51 categories. }

\begin{figure}
    \centering
    \includegraphics[width=1\linewidth]{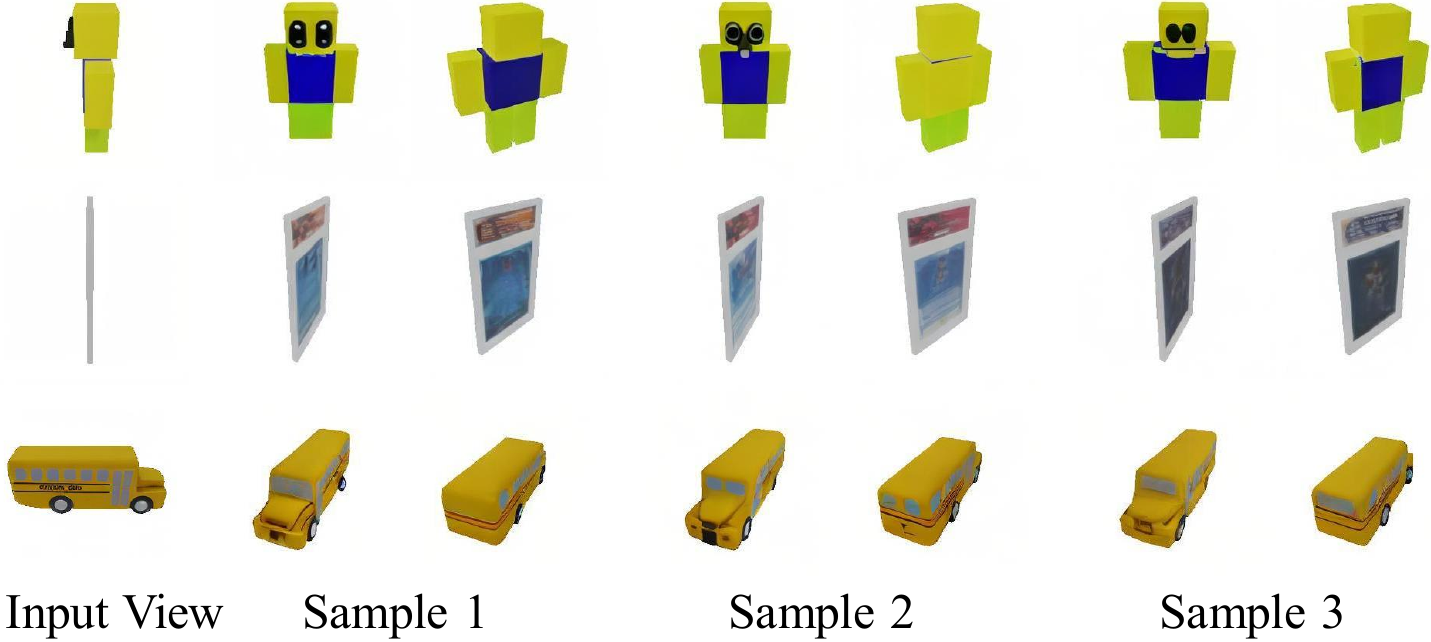}
    \caption{\textbf{Sample Diversity.} \ours~is capable of generating diverse samples given the same input. We show the input image (left) followed by views synthesized in three randomly generated samples. We observe that there is meaningful variation in uncertain regions \eg the eyes of the character and the colors on the screen vary across samples.}
    \figlabel{nvs-gso-diversity}
\end{figure}

\vspace{-2mm}
\paragraph{Baselines.}
For the novel view synthesis task, we adopt Zero-1-to-3~\cite{liu2023zero1to3} and SyncDreamer~\cite{liu2023syncdreamer} as our baseline methods. Given an input image, Zero-1-to-3 can synthesize images from novel viewpoints. Built on Zero-1-to-3, SyncDreamer can simultaneously generate multiple images from different viewpoints with 3D consistency. 
\camready{For CO3D, we compare against PixelNeRF \cite{yu2021pixelnerf} as both Zero-1-to-3 and SyncDreamer are restricted to 3-DoF camera variation. }

For 3D reconstruction, we compare our method with the aforementioned two methods together with RealFusion~\cite{melas2023realfusion}, Magic 123~\cite{qian2023magic123}, One-2-3-45~\cite{liu2023one}, Point-E~\cite{nichol2022point} and Shape-E~\cite{jun2023shap}. Note that the diffusion-based methods require neural field optimization using either rendering objectives or distillation objectives (\eg Zero-1-to-3 requires a SDS distillation for extracting geometry whereas SyncDreamer relies on training  Neus~\cite{wang2021neus}), whereas our method allows `directly' computing the geometry via un-projecting the predicted depth maps. To highlight this distinction, we categorize the reconstruction approaches as direct (One-2-3-45~\cite{liu2023one}, Point-E~\cite{nichol2022point}, Shape-E~\cite{jun2023shap}, and \ours) or optimization based (RealFusion~\cite{melas2023realfusion}, Magic 123~\cite{qian2023magic123}, Zero-1-to-3~\cite{liu2023zero1to3}, and SyncDreamer~\cite{liu2023syncdreamer}).

\vspace{-2mm}
\paragraph{Metrics.}
For the novel view synthesis, we adopt commonly used metrics: PSNR, SSIM~\cite{wang2004image}, and LPIPS~\cite{zhang2018unreasonable}. For the 3D reconstruction task, we report Chamfer Distances between ground-truth and predicted point clouds. 

\vspace{-2mm}
\paragraph{Implementation Details.}
We train our model on a filtered version of the Objaverse dataset, which consists of about 400k instances. During training, for each instance, we randomly sample 5 views and choose the first one as the input view. We train the model for 400k iterations with 4 40G A100 GPUs using a total batch size of 16. We use Adam optimizer with a learning rate of 1e-5. Even though we only train with 5 views, we can sample an arbitrary set during inference as our depth-based projection followed by transformer-based aggregation trivially generalizes to incorporate more views. In our experiments, we render 16 views for each instance for evaluation.

\subsection{Novel View Synthesis}
\seclabel{exp_obja}

\begin{table}[t]
\small
\renewcommand\arraystretch{0.9}
\caption{\textbf{Results for novel view synthesis on the Objaverse dataset}. We compare our method with two other baselines on 100 instances from the test set of Objaverse dataset, same as in \cite{liu2023zero1to3}. Our method outperforms existing baselines over three commonly used metrics: PSNR, SSIM and LPIPS.}

\begin{tabularx}{8.2cm}{p{2.5cm}X<{\centering}X<{\centering}X<{\centering}}

\toprule[1.5pt]
Method  & PSNR $\uparrow$ & SSIM $\uparrow$ & LPIPS  $\downarrow$\\
\midrule[1pt]
Zero123~\cite{liu2023zero1to3} &       17.37 & 0.783 & 0.211  \\
SyncDreamer~\cite{liu2023syncdreamer}  &  \second 19.22 & \second 0.817 & \second 0.176 \\
 \ours  &        \first 21.19 & \first 0.835 & \first 0.146 \\
\bottomrule[1.5pt]
\end{tabularx}

\vspace{-2mm}

\tablelabel{nvs-obja}
\end{table}

\begin{table}[t]
\small
\renewcommand\arraystretch{0.9}
\caption{\textbf{Results for novel view synthesis on the Google Scanned Objects (GSO) dataset.} We compare our method with two other baselines on 100 instances randomly chosen from the GSO dataset. Our method achieves consistent improvement over baseline methods on PSNR and LPIPS, while slightly worse than SyncDreamer on SSIM.}
\begin{tabularx}{8.2cm}{p{2.5cm}X<{\centering}X<{\centering}X<{\centering}}
\toprule[1.5pt]
Method  & PSNR $\uparrow$ & SSIM $\uparrow$ & LPIPS $\downarrow$  \\
\midrule[1pt]
Zero123~\cite{liu2023zero1to3}         & 17.42 & 0.756 & 0.207\\
SyncDreamer~\cite{liu2023syncdreamer}     & \second 18.95 &  \first 0.796 & \second 0.176 \\
 \ours           & \first 19.53 & \second 0.790 & \first 0.175 \\
\bottomrule[1.5pt]
\end{tabularx}

\tablelabel{nvs-gso}
\end{table}

\begin{table}[t]
\small
\renewcommand\arraystretch{0.9}
\caption{\textbf{Results for 51 category novel view synthesis on CO3D.} We significantly outperform PixelNeRF on perceptual quality. }
\begin{tabularx}{8.2cm}{p{2.5cm}X<{\centering}X<{\centering}X<{\centering}}
\toprule[1.5pt]
Method  & PSNR $\uparrow$ & SSIM $\uparrow$ & LPIPS $\downarrow$  \\
\midrule[1pt]
PixelNeRF~\cite{yu2021pixelnerf}    & \first 17.64    & \second 0.484  & \second 0.378\\
 \ours                              & \second 17.16   & \first 0.701 & \first 0.220 \\
\bottomrule[1.5pt]
\end{tabularx}

\tablelabel{nvs-co3d}
\end{table}

\begin{figure}[h]
    \vspace{-0em}
    \centering
    \includegraphics[width=1\linewidth]{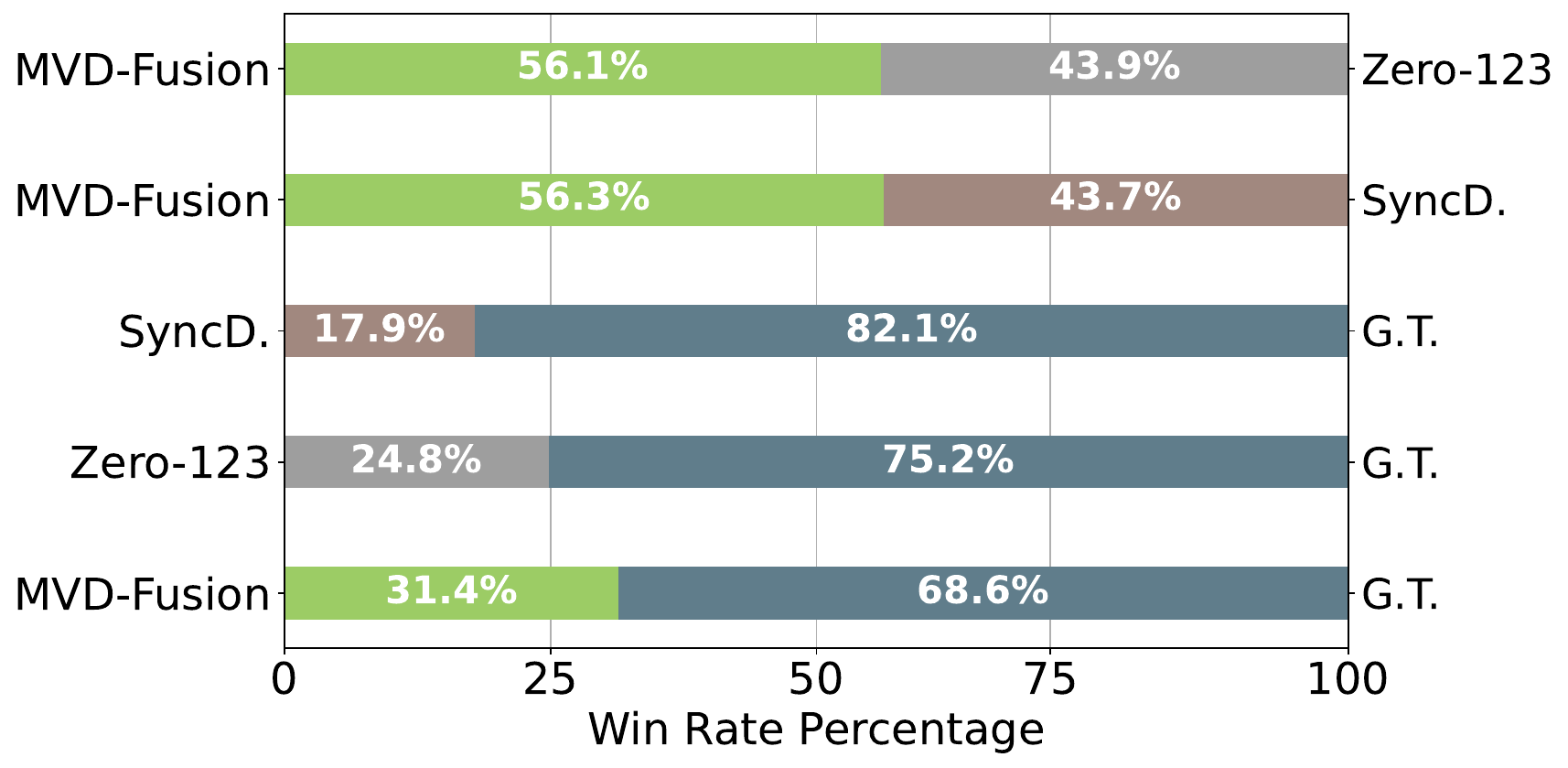}
    \vspace{-2em}
    \caption{User preference percentage of \ours~against Zero-123, SyncDreamer (SyncD.), and ground truth (G.T.).}
    \figlabel{user-study}
\end{figure}

\paragraph{Objaverse and Google Scanned Objects.} We report quantitative results on the Objaverse dataset and GSO dataset in \tableref{nvs-obja} and \tableref{nvs-gso}, respectively. For the Objaverse dataset, we use the held-out test set for evaluation whereas we use a subset of 30 random objects from GSO. We find that our method achieves consistent improvements over the baselines across metrics on both, the in-distribution Objaverse dataset and the out-of-distribution GSO dataset. We also provide qualitative comparisons on the Objaverse dataset and GSO dataset in \figref{nvs-obja-gso-fig}. Although Zero-1-to-3~\cite{liu2023zero1to3} produces visually reasonable images, it suffers from multi-view inconsistency across generated viewpoints. In contrast SyncDreamer and \ours, are able to obtain multi-view consistency among generated images. However, SyncDreamer struggles to obtain high alignment with the input image, sometimes leading to consistent multi-view images with unreasonable appearance. Our method, on the other hand, is able to generate multi-view consistent images with better alignment with the image input and more plausible completions in unobserved regions. Moreover, we also note that SyncDreamer is trained on the whole Objaverse dataset and may have actually seen these instances whereas these represent held out instances for both, \ours~ and Zero-1-to-3.

In addition to visualizing the comparative results with baselines, we also highlight the ability of \ours~ to generate multiple plausible outputs. In particular, since novel view synthesis from a single image is an under-constrained task, using the diffusion model can effectively generate more diverse samples given a single input image. As shown in \figref{nvs-gso-diversity}, \ours~ is able to generate diverse plausible samples with different random seeds \eg varying textures in the front of the bus.

\vspace{-2mm}
\paragraph{User Study.} We run a user study by randomly selecting 40 instances from Objaverse and GSO test set and asking 43 users to make 860 pairwise comparisons (users are shown an input image and two generated novel views per method). We show results in \figref{user-study}. Our method tends to be chosen over Zero-123 and SyncDreamer, and is also more competitive with GT compared to these baselines.

\vspace{-2mm}
\paragraph{Common Objects in 3D.} \camready{Real-world data often have cameras that are not origin facing, making methods that model 3DoF origin facing cameras \cite{liu2023zero1to3, liu2023syncdreamer} not suitable for real-world inference. We finetune our model on CO3D and show novel view synthesis results in \tableref{nvs-co3d}. We also train a cross-category PixelNeRF~\cite{yu2021pixelnerf} model as a baseline. While it  is slightly better in PSNR (perhaps due to blurry mean predictions being optimal under uncertainty), our method vastly outperforms PixelNeRF in perceptual metrics SSIM and LPIPS (see \figref{co3d}). }

\begin{figure}
    \centering
    \includegraphics[width=1\linewidth]{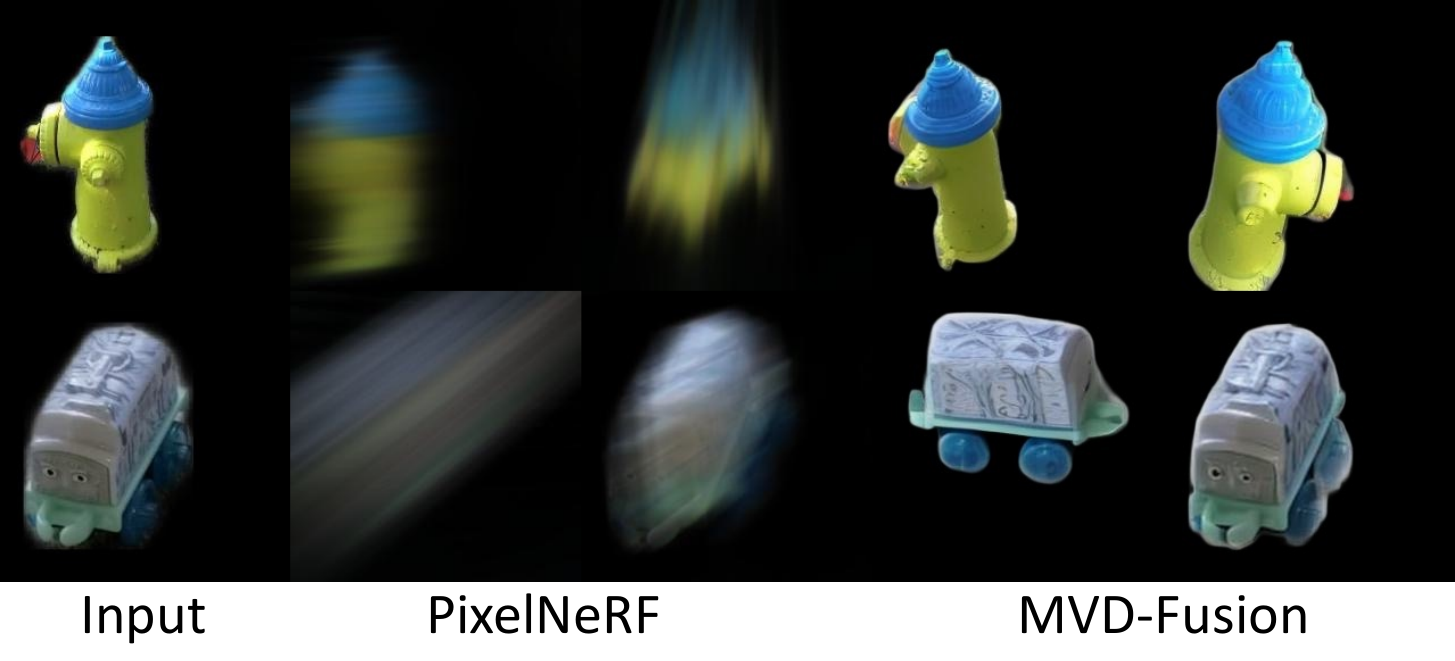}
    \vspace{-2em}
    \caption{Qualitative results for novel view synthesis on instances from CO3D. \ours is able to predict accurate and realistic novel views on real-world datasets with perspective camera poses. }
    \figlabel{co3d}
    \vspace{-1.5em}
\end{figure}

\begin{figure*}
    \centering
    \includegraphics[width=1\linewidth]{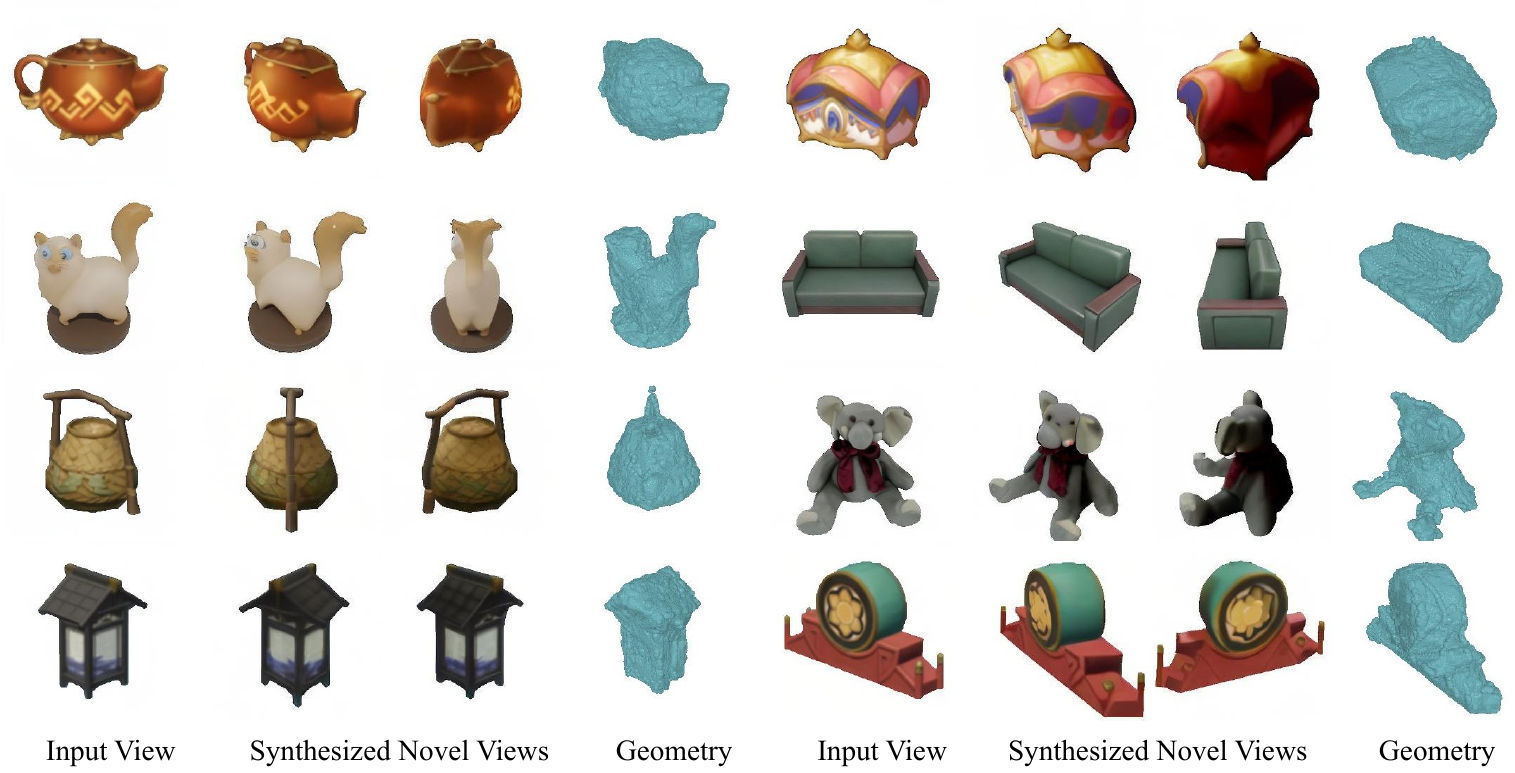}
    \caption{\textbf{In-the-wild Generalization.} We visualize the prediction from \ours~on in-the-wild internet images. We find that \ours~ is able to preserve the rich texture in the input images and model the rough geometry without post-processing.}
    \figlabel{nvs-wild}
    \vspace{-1.1em}
\end{figure*}
\subsection{Single-view Reconstruction} 
\seclabel{exp_3d}

\begin{table}[t]
\renewcommand\arraystretch{0.9}
\caption{\textbf{Results for 3D reconstruction on the Google Scanned Objects (GSO) dataset.}  `Optimization' denotes methods that require additional training such as fitting an occupancy field to obtain 3D shapes. `Direct' denotes methods that can directly output 3D predictions. Following ~\cite{liu2023syncdreamer}, we report Chamfer Distance on the same 30 instances from the GSO dataset. Our method demonstrates consistent improvement over `direct' methods and outperforms most of the `optimization' methods. }
\vspace{-1em}
\begin{center}
\begin{tabularx}{8.2cm}{p{2.0cm}p{2.7cm}X<{\centering}X<{\centering}}
\toprule[1.5pt]
3D Extraction & Method  & Chamfer Dist $\downarrow$ \\
\midrule[1pt]
\multirow{4}{*}{Optimization}
    & RealFusion~\cite{melas2023realfusion}    & 0.082 \\
    & Magic123~\cite{qian2023magic123}      & 0.052 \\
    & Zero123~\cite{liu2023zero1to3}       & 0.034  \\
    & SyncDreamer~\cite{liu2023syncdreamer}   & \first 0.026 \\
\midrule
\multirow{4}{*}{Direct}
    & One-2-3-45~\cite{liu2023one}    & 0.063 \\
    & Point-E~\cite{nichol2022point}      & 0.043 \\
    & Shap-E~\cite{jun2023shap}       & 0.044 \\
    & \ours          & \second 0.031   \\
\bottomrule[1.5pt]
\end{tabularx}
\end{center}
\tablelabel{3d-gso}
\vspace{-2em}
\end{table}

Unlike previous methods such as Zero-1-to-3 and SyncDreamer, which have to fit a radiance field from generated multi-view images to obtain the 3D shape, \ours~can directly obtain the point cloud. With multi-view RGB-D generations, we can simply unproject the foreground pixels and obtain the object point cloud. We show quantitative results in \tableref{3d-gso}, where we compare our method against previous methods on the GSO dataset using chamfer distance. We see that our method outperforms all of the methods that directly infer 3D shapes and most of the methods that require further optimization steps to get the 3D shapes. 

\subsection{In-the-wild Generalization}
\seclabel{exp_add}
We also demonstrate the generalization ability of \ours~for reconstructing in-the-wild images from the internet. We show qualitative results depicting generated novel views and recovered point clouds in \figref{nvs-wild}. With challenging out-of-domain images as input, \ours~is still capable of generating consistent novel-view images and reasonable 3D shapes from single-view observation.


\section{Discussion}
In this work, we presented \ours, which allowed co-generating multi-view images given a single input image. Our approach allowed adapting a pre-trained large-scale novel-view diffusion model for generating multi-view RGB-D images, and enforced consistency among these via depth-guided projection. While our results showed improvements over prior state-of-the-art across datasets, there are several challenges that still remain. First, the multi-view consistency is encouraged via inductive biases in the network design but is not guaranteed, and the network may generate (slightly) inconsistent multi-view predictions. Moreover,  while our inferred multi-view depth maps can yield a point cloud representation that captures the coarse geometry, these coarse depth maps do not capture the fine details visible in the generated views and an optimization-based procedure may help extract these better. Finally,  our approach has been trained on clean unoccluded instances and would not be directly applicable under cluttered scenes with partially visible objects, and it remains an open research question to build systems that can deal with such challenging scenarios.

\section*{Acknowledgements}
We thank Bharath Raj, Jason Y. Zhang, Yufei (Judy) Ye, Yanbo Xu, and Zifan Shi for helpful discussions and feedback. This work is supported in part by NSF GRFP Grant No. (DGE1745016, DGE2140739).

\newpage
{
    \small
    \bibliographystyle{ieeenat_fullname}
    \bibliography{main}
}

\clearpage
\setcounter{page}{1}
\maketitle
\appendix


\section{Architecture Details}
\seclabel{supp_arch}
Our network consists of modifications on top of Zero123 \cite{liu2023zero1to3}. We describe each component of our network in detail. 

\parnobf{VAE.}
We use the pretrained VAE from Stable Diffusion 1.4 \cite{rombach2022high}. We freeze the VAE. 

\parnobf{UNet.}
We initialize our UNet with weights from Zero123 \cite{liu2023zero1to3}. Zero123 has a novel view synthesis UNet that accepts one input image (4 channel latents) and one target noisy image (4 channel latents) along with camera pose and predicts a novel view image latent (4 channels). We modify the input and output blocks to accommodate the prediction of an additional depth channel. Our UNet has 10 input channels and 5 output channels. For all experiments, we use only RGB images as input (4 channel latents) and pad the additional channel with zeros. The noisy target target image is always 5 channels. 

\parnobf{CLIP and Camera Pose Embedding.} We follow Zero123 \cite{liu2023zero1to3} to use the frozen CLIP \cite{radford2021clip} image encoder along with camera information as one of the inputs to the cross-attention layers in Stable Diffusion. However, instead of using azimuth and elevation angle representation, we directly use a flattened camera matrix as input. We use 3 fully connected layers to map CLIP image embedding and flattened camera matrix into cross-attention input of dimension 768.

\parnobf{Depth-guided Multi-view Attention.}
After each of the existing cross-attention layers in the UNet, we add additional cross attention layers that attend to view-aligned feature frustums sampled from our depth-guided multi-view attention module. Our depth-guided attention module is a 3 layers transformer that aggregates information across the noisy target latents from the current timestep and also input image latents. For each target view, we generated a feature frustum of shape (1, 256, 3, 32, 32), where the feature map is 32 by 32, with 3 depth samples and feature dimension 256. The depth dimension represents the number of depth points sampled along each ray and can be reduced down to just 1. Our transformer uses a hidden dimension of 256 with 8 heads. We use an additional fully connected layer to project our features into 768 dimensions, making them compatible with existing cross attention layers. A key difference between our multi-view cross attention and text cross attention is that, in our multi-view attention, each latent patch independently attends to the corresponding patch in the feature frustum. 





\section{Additional Results and Visualizations}
\seclabel{supp_visual}
\parnobf{Ablating Classifier-free Guidance Scale.} 
\begin{table}[t]
\renewcommand\arraystretch{0.9}

\caption{Ablation study on Google Scanned Objects (GSO) dataset. We ablate the effect of the classifier-free guidance scale during inference. We randomly chose 30 instances from the dataset for evaluation. `Scale' denotes the classifier-free guidance scale.}
\begin{tabularx}{8.2cm}{p{1.5cm}X<{\centering}X<{\centering}X<{\centering}}
\toprule[1.5pt]
Scale  & PSNR $\uparrow$ & SSIM $\uparrow$ & LPIPS  $\downarrow$\\
\midrule[1pt]
1.0       & 19.34 & 0.775 & 0.199 \\
2.0     & \first 19.91 & \first 0.787 & \first 0.184 \\
3.0     & 19.38 & 0.778 & 0.189 \\
5.0     &  18.66   & 0.771 & 0.194 \\
\bottomrule[1.5pt]
\end{tabularx}

\tablelabel{ablation}
\end{table}
We further conduct experiments to ablate the effect of the classifier-free guidance scale. Proposed in ~\cite{ho2022classifier}, classifier-free guidance controls the faithfulness of the generated output to the conditional input. In our method, we use the classifier-free guidance scale $\omega$ to control the contribution of the classifier-free model: $\hat{\epsilon}_{\phi'}(\mathbf{y}, \mathbf{x}_t^n, \mathbf{\pi}^n, \mathbf{z}_t^n, t) = \omega \epsilon_{\phi'}(\mathbf{y}, \mathbf{x}_t^n, \mathbf{\pi}^n, \mathbf{z}_t^n, t) + (1-\omega)\epsilon_{\phi'}(\mathbf{x}_t^n, t)$, where $\epsilon_{\phi'}(\mathbf{y}, \mathbf{x}_t^n, \mathbf{\pi}^n, \mathbf{z}_t^n, t)$ is our proposed multi-view diffusion model. In practice, we notice that a higher classifier-free guidance scale leads to better multi-view consistency. As shown in \tableref{ablation}, we find that adopting a scale of 2 yields the best performance. Therefore, we use this classifier-free guidance scale for inference.






\end{document}